\documentclass[acmtog]{acmart}

\usepackage{makecell}
\usepackage{pifont}

\usepackage[table]{xcolor}
\usepackage{stfloats}
\usepackage{float}
\usepackage{xspace}

\newcommand{\ourmethod}{CHIP\xspace} 

\copyrightyear{2026}
\acmYear{2026}
\setcopyright{cc}
\setcctype{by-nc-nd}
\acmConference[SA Conference Papers '26]{SIGGRAPH Asia 2026 Conference Papers}{December 01--04, 2026}{Kuala Lumpur, Malaysia}
\acmBooktitle{SIGGRAPH Asia 2026 Conference Papers (SA Conference Papers '26), December 01--04, 2026, Kuala Lumpur, Malaysia}
\acmDOI{10.1145/3829340.3842357}
\acmISBN{979-8-4007-2842-6/2026/12}

\begin{document}
\title{Unifying Physics-Based Humanoid Interaction with a Context-Conditioned Interaction Prior}

\author{Jianan Li}
\email{jnli22@cse.cuhk.edu.hk}
\affiliation{%
  \institution{The Chinese University of Hong Kong}
  \department{Department of Computer Science and Engineering}
  \city{Hong Kong}
  \country{China}
}

\author{Xiao Chen}
\email{cx123@ie.cuhk.edu.hk}
\affiliation{%
  \institution{The Chinese University of Hong Kong}
  \department{Department of Information Engineering}
  \city{Hong Kong}
  \country{China}
}

\author{Tien-Tsin Wong}
\email{tt.wong@monash.edu}
\affiliation{%
  \institution{Monash University}
  \department{Department of Data Science and AI}
  \city{Clayton}
  \country{Australia}
}

\begin{abstract}
Developing unified physics-based humanoid controllers that can navigate complex 3D scenes and manipulate objects remains a longstanding challenge. Existing approaches are often specialized for either locomotion or object-centric manipulation, or rely on task-specific reward engineering that does not scale well across diverse behaviors. We present \ourmethod, a unified, physics-grounded framework for learning reusable humanoid interaction skills from heterogeneous motion data. Central to our approach is a conditional interaction prior that models a context-dependent distribution over these skills within a shared discrete space. Our method is trained in three stages. We first learn physics-based motion-imitation policies that acquire grounded teacher behaviors from heterogeneous interaction data. We then distill these behaviors into a context-conditioned interaction prior that captures reusable motion structure across locomotion and manipulation. Finally, we initialize downstream task policies from the pretrained prior and adapt them through prior-regularized online RL post-training. Experiments on a diverse suite of humanoid interaction tasks show that our approach supports scene-aware locomotion, contact-rich object manipulation, and compositional behaviors such as environment-aware object transport and long-horizon skill sequencing, while producing smooth transitions and physically plausible motion. Project page: \url{https://jiann-li.github.io/chip-project/}.
\end{abstract}



\begin{teaserfigure}
  \centering
  \includegraphics[width=\textwidth]{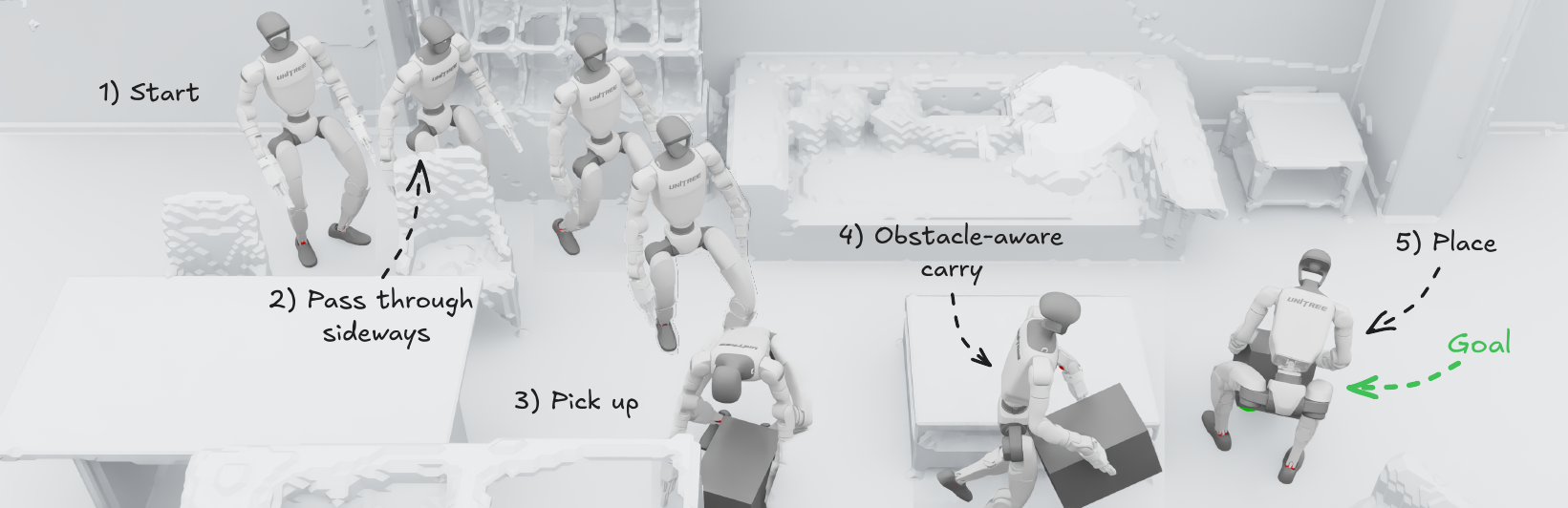}
  \caption{\textbf{\ourmethod enables multi-stage humanoid interaction in complex 3D environments.} Representative rollouts demonstrate whole-body collision avoidance when passing sideways through narrow gaps, contact-rich object manipulation, and compositional capabilities such as obstacle-aware object carrying within a single continuous behavior.}
  \Description{Teaser figure showing qualitative examples of multi-stage humanoid interaction behaviors, where the humanoid composes navigation and manipulation skills across multiple stages of a task.}
  \label{fig:teaser}
\end{teaserfigure}
\maketitle
\section{Introduction}
Controlling physics-based humanoids that interact naturally with complex 3D environments remains a long-standing challenge in computer graphics, character animation, robotics, and embodied intelligence.
Real-world humanoid behavior is inherently context-dependent: human motions continuously adapt to scene geometry, object state, and task goals.
For example, a humanoid may need to squeeze through narrow gaps in cluttered spaces, adjust its whole-body posture to avoid obstacles, and then coordinate contact-rich motion to move an object toward a target location.
The challenge lies not only in acquiring a diverse repertoire of motor skills but also in synthesizing robust behaviors that simultaneously satisfy high-level goals, spatial constraints, and physical laws.

Unifying these highly context-dependent behaviors into a shared physics-based framework exposes the limitations of existing approaches. Reinforcement learning (RL) controllers, including recent unified approaches, still typically acquire supported behaviors through task-specific objectives and supervision, requiring additional reward design and policy optimization as new skills are introduced~\cite{hassan2023synthesizing, pan2024synthesizing, pan2025tokenhsi}. To improve scalability, recent methods distill diverse skills acquired through large-scale motion tracking into reusable generative priors.
However, these priors remain largely environment-agnostic, without explicitly grounding skill selection in scene constraints and affordances~\cite{peng2022ase, luouniversal, zhu2023neural, tessler2024maskedmimic, truong2024pdp, huang2025diffuse}.
Even methods that incorporate object manipulation typically model object interactions independently of broader scene context, without capturing their coupling with full-body navigation~\cite{tessler2025maskedmanipulator, xu2026interprior}.
While some recent kinematics-based approaches begin to co-model scene and object interactions~\cite{liu2025uni}, they are not grounded in physical simulation. Consequently, developing a unified, physics-grounded framework that explicitly conditions humanoid skills on heterogeneous interaction contexts remains an open challenge.

In this work, we introduce \textbf{\ourmethod} (\textbf{C}ontext-conditioned \textbf{H}umanoid \textbf{I}nteraction \textbf{P}rior), a unified, physics-grounded control framework for humanoid agents in complex 3D environments.
The key insight is to organize diverse humanoid behaviors around interaction context rather than motion alone.
To this end, \ourmethod learns a context-conditioned generative prior that models the distribution of reusable interaction skills conditioned on humanoid state and scene, object, and task context. 
A shared conditioning interface enables human-scene interaction (HSI) and human-object interaction (HOI) behaviors with different spatial and task specifications to be distilled into the same interaction prior.
We train \ourmethod in three stages: physics-based motion imitation first acquires grounded teacher behaviors, generative distillation then compresses these behaviors into a shared discrete skill prior, and prior-regularized RL post-training adapts the prior to downstream tasks.

We evaluate \ourmethod on a diverse suite of humanoid interaction tasks in complex 3D environments.
Experimental results show that policies adapted from the shared interaction prior support scene-aware locomotion, contact-rich object manipulation, and compositional behaviors such as environment-aware object transport. Qualitative results further demonstrate continuous transitions between locomotion and manipulation. Our main contributions are as follows:
\begin{itemize}
    \item A physics-grounded control framework that integrates scene-level navigation and object-level manipulation through a shared interaction prior.
    \item A context-conditioned categorical interaction prior that grounds discrete skill distributions in heterogeneous scene, object, and task context, enabling reusable skill modeling beyond motion alone.
    \item A three-stage learning strategy that separates physics-based behavior acquisition, context-conditioned prior distillation, and prior-regularized downstream adaptation.
\end{itemize}
\section{Related Work}
\subsection{Physics-based Character Animation}\label{sec:chara_control}
Physics-based character animation is predominantly learned through reinforcement learning in simulation. Early work relied heavily on expert-designed reward functions to acquire locomotion and agile motor skills~\cite{schulman2015high, peng2017deeploco, xie2020allsteps, lee2010data, coros2010generalized, tao2022getup, peng2016terrain, liu2016guided, liu2012terrain, yin2021discovering}. While effective on well-specified tasks, such approaches often require substantial task-specific engineering and are difficult to scale to diverse behaviors.

To reduce manual reward design, later work increasingly incorporated motion capture data into policy learning to improve the naturalness and diversity of synthesized motions. One important line of research uses adversarial imitation learning to match the distribution of generated behaviors to motion data while simultaneously optimizing task objectives~\cite{peng2021amp, hassan2023synthesizing, gao2024coohoi, juravsky2022padl, tessler2023calm, wang2025sims}. Although successful in producing natural motions, adversarial approaches often suffer from unstable training and reduced behavioral diversity due to mode collapse.

Another influential paradigm is motion tracking, in which a character is trained to reconstruct reference motions directly in simulation~\cite{peng2018deepmimic, fussell2021supertrack, li2025learning}. Compared with adversarial imitation, motion tracking provides a simple and scalable recipe for learning large repertoires of skills from motion datasets~\cite{luo2023perpetual}, and has also been extended beyond locomotion to richer interaction settings~\cite{wang2023physhoi, xu2025intermimic, yu2025skillmimic}. However, tracking-based controllers require accurate reference trajectories as input, which limits their flexibility for downstream control and task adaptation.

More recently, data-driven behavior priors have improved the scalability of physics-based character animation by learning reusable skill representations that can be coupled with hierarchical controllers for downstream tasks~\cite{peng2022ase,luouniversal, yao2022controlvae, yao2024moconvq, zhu2023neural, luo2024omnigrasp}. However, these priors are often task-agnostic and primarily capture motion-level dynamics, leaving downstream controllers to rely on reward design for task-specific objectives. This makes it difficult to efficiently learn complex interaction behaviors that must adapt to scene geometry or object state.

\subsection{Unified Humanoid Controllers}
To improve scalability beyond behavior-specific policies, recent work has explored unified humanoid controllers that consolidate multiple skills within a shared policy or generative model. One line of work designs universal task interfaces that express locomotion and interaction tasks within a unified goal-conditioned formulation~\cite{xiao2024unified, deng2026humanobject, pan2025tokenhsi}. These methods enable multi-task execution, but their flexibility is often limited by predefined task interfaces, making it difficult to adapt to novel capabilities or interaction modes. Another line of work learns in-painting-based generative controllers that synthesize natural and physically plausible humanoid behaviors from sparse kinematic constraints~\cite{tessler2024maskedmimic, tessler2025maskedmanipulator, wu2025uniphys, huang2025diffuse, xu2026interprior}.

While these approaches point toward generalist humanoid control, they still fall short for rich interaction in complex 3D environments. Task-interface methods are constrained by predefined specifications, while in-painting-based controllers often lack explicit modeling of scene geometry and object-level constraints. In contrast, our work learns a physics-grounded interaction prior from heterogeneous HSI and HOI data with a shared scene- and object-aware conditioning interface, enabling the learned skills to be reused, adapted, and composed in complex 3D environments.

\section{Method}\label{sec:method}
\begin{figure*}[t]
    \centering
    \includegraphics[width=\textwidth]{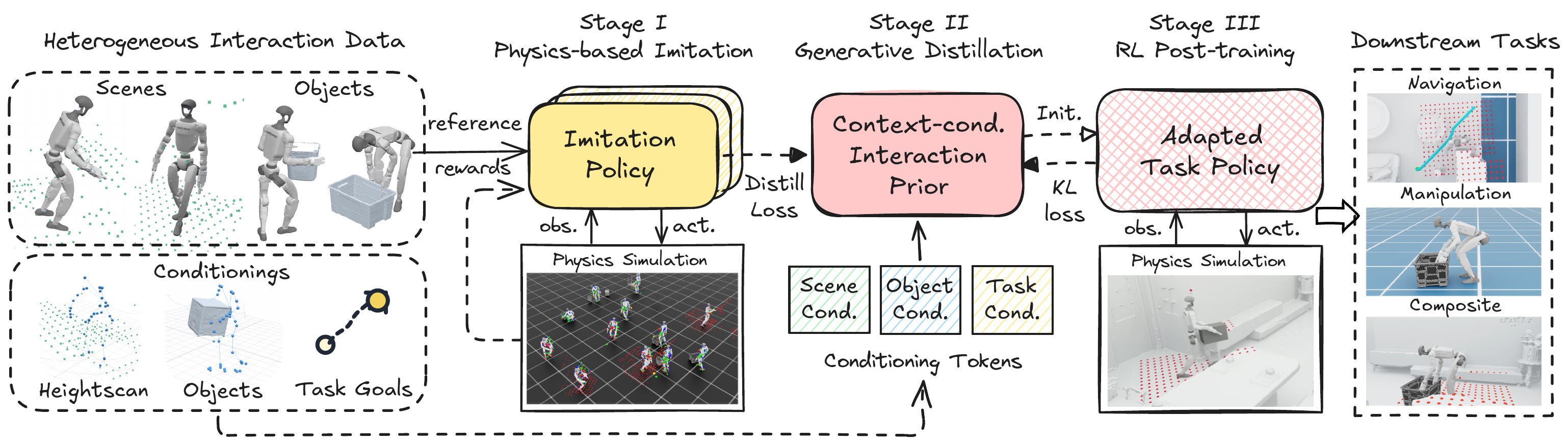}
    
    \caption{\textbf{Overview of the proposed framework.} 
    Starting from heterogeneous HSI and HOI motion data, Stage I trains physics-based tracking policies to reproduce scene- and object-interaction behaviors. 
    Stage II distills their behaviors into a shared discrete skill representation and a context-conditioned interaction prior using multimodal scene, object, and task inputs.
    Stage III initializes a downstream task policy from the pretrained prior and adapts the policy with reinforcement learning, while KL regularization toward the frozen prior preserves the learned interaction structure and motion quality.}
    
    \Description{Left-to-right overview of the three-stage framework. Heterogeneous scene- and object-interaction motions and their conditioning signals feed physics-based imitation experts. Their behaviors are distilled into a context-conditioned generative skill prior, which initializes an RL task policy regularized toward the frozen prior for navigation, manipulation, and composite tasks.}
    \label{fig:main_pipeline}
\end{figure*}

\subsection{Overview}
We propose a unified, physics-grounded framework for learning reusable skills across environment-aware locomotion, object-centric manipulation, and their composition.
Our key insight is to organize reusable skills around interaction context rather than motion alone. We distill heterogeneous HSI and HOI behaviors into a shared discrete interaction prior whose skill distribution is conditioned on scene, object, and task context.
As illustrated in Figure~\ref{fig:main_pipeline}, we first train physics-based imitation experts, then distill them into a context-conditioned generative prior, and finally adapt it to downstream tasks through prior-regularized RL post-training.

\subsection{Problem Formulation}
We model humanoid interaction as a context-augmented, goal-conditioned Markov decision process (MDP). We denote the MDP by $\mathcal{M}=(\mathcal{S},\mathcal{A},p,r,\mathcal{G},\mathcal{C})$. Here, $\mathcal{S}$ and $\mathcal{A}$ denote the proprioceptive state and action spaces, $p$ the simulator dynamics, $r$ the task reward, $\mathcal{G}$ the goal space, and $\mathcal{C}$ the multimodal interaction-context space capturing scene and object constraints. At step $t$, the agent observes $s_t \in \mathcal{S}$, $g_t \in \mathcal{G}$, and $c_t \in \mathcal{C}$, and executes $a_t \sim \pi(a_t \mid s_t,c_t,g_t)$.

We assume access to a heterogeneous interaction dataset $\mathcal{D}=\{(\hat{\xi}^j,c^j,g^j)\}_{j=1}^{M}$ containing reference motions $\hat{\xi}^j$, interaction contexts $c^j$, and goals $g^j$. Since $\mathcal{D}$ combines heterogeneous HSI and HOI data, individual sequences may contain only a subset of the context modalities. Our objective is to leverage $\mathcal{D}$ to learn a reusable interaction prior over latent skills, which subsequently initializes and regularizes a downstream task policy for maximizing the expected discounted return
$J(\pi) = \mathbb{E} \big[ \sum_{t} \gamma^t r(s_t, a_t) \big]$.

\subsection{Stage I: Motion Imitation from Heterogeneous Data}
To acquire physically grounded teacher behaviors for subsequent generative distillation, we perform physics-based motion imitation over a diverse corpus of humanoid behaviors, encompassing both human-scene interactions and human-object interactions.

\paragraph{Observations.}
At each step $t$, the motion imitation policy's observation $o_t^E$ consists of the humanoid's proprioceptive state and future reference kinematics over a finite horizon:
\begin{equation}
o_t^E = (s_t^{\mathrm{prop}}, \hat{\xi}_{t:t+L}, \hat{\xi}^\mathrm{obj}_{t:t+L})
\end{equation}
The proprioceptive state $s_t^{\mathrm{prop}}$ encodes the humanoid's base orientation, linear and angular velocities, and joint positions and velocities. The reference features $\hat{\xi}_{t:t+L}$ encode the target humanoid kinematics over a future horizon $L$, represented relative to the current humanoid state in terms of position, orientation, and velocity. For HOI sequences, we additionally include the corresponding future object kinematics $\hat{\xi}_{t:t+L}^{\mathrm{obj}}$.

\paragraph{Rewards.}
The motion imitation policies are trained with a tracking reward
\begin{equation}
    r_t^E = \lambda_p r_t^{\text{pos}} + \lambda_q r_t^{\text{rot}} + \lambda_v r_t^{\text{vel}} + \lambda_\omega r_t^{\text{ang}} + \lambda_o r_t^{\text{obj}} + \lambda_r r_t^{\text{reg}},
\end{equation}
where $r_t^{\text{pos}}$, $r_t^{\text{rot}}$, $r_t^{\text{vel}}$, and $r_t^{\text{ang}}$ measure tracking consistency in position, rotation, linear velocity, and angular velocity, respectively.
The position and rotation terms combine pelvis-anchor and aligned body-pose tracking, while the velocity terms track global body linear and angular velocities. For HOI experts, $r_t^{\text{obj}}$ combines object relative-position and global-orientation tracking and is set to zero otherwise. The regularization term consists of action-rate, joint-limit, and undesired-contact penalties. Full reward definitions and weights are provided in the supplement.

\paragraph{Training.}
We train specialist experts for different interaction categories, including locomotion and object interactions, using reference-state initialization and kinematic early termination~\cite{peng2018deepmimic}. This stage focuses on robust behavior acquisition; heterogeneous experts are subsequently unified through the Stage-II generative distillation.

\subsection{Stage II: Distilling the Unified Interaction Prior}
Using the physics-based imitation policies as teachers, we next distill their behaviors into a shared discrete skill representation and a context-conditioned interaction prior. This stage consolidates physically grounded HSI and HOI behaviors within a common latent space while conditioning skill selection on interaction contexts and task commands. Concretely, a vector-quantized variational autoencoder (VQ-VAE) compresses teacher behaviors into discrete skill tokens, and a transformer-based categorical prior models their conditional distribution.

\paragraph{Unified Skill Representation.}
To obtain a compact and reusable representation of heterogeneous interaction behaviors, we learn a shared discrete skill space using a VQ-VAE~\cite{van2017neural}. At each time step $t$, the encoder maps the motion-imitation observation $o_t^E$ to a continuous latent $z_t$, which is quantized against a learned codebook $\mathcal{Z}=\{z^{(k)}\}_{k=1}^{K}$ to obtain a discrete skill token $q_t$. The corresponding codebook embedding is decoded into an action distribution conditioned on the humanoid proprioceptive state $s_t^{\mathrm{prop}}$:
\begin{equation}
\begin{aligned}
\text{Encoder:} \quad z_t &= E_{\eta}(o_t^E), \\
\text{Quantizer:} \quad q_t &= Q(z_t), \quad q_t \in \{1,\dots,K\}, \\
\text{Decoder:} \quad a_t &\sim D_{\psi}(\cdot \mid s_t^{\mathrm{prop}}, z^{(q_t)}).
\end{aligned}
\end{equation}
Here, $z^{(q_t)}$ denotes the codebook embedding associated with token $q_t$. The discrete skill space captures reusable motor behaviors, while the decoder translates the selected skill token into joint-space proportional--derivative (PD) targets.
Using teacher actions $a_t^{E}$, we combine the behavior-cloning objective $\mathcal{L}_{\mathrm{bc}}$ with a commitment loss:
\begin{equation}
\mathcal{L}_{\mathrm{VQ}}=\mathcal{L}_{\mathrm{bc}}+\beta_{\mathrm{com}}\sum_t\|z_t-\operatorname{sg}[z^{(q_t)}]\|_2^2.
\end{equation}
Here, $\mathcal{L}_{\mathrm{bc}}$ is the negative log-likelihood of the teacher actions under $D_{\psi}$, and $\operatorname{sg}[\cdot]$ denotes stop-gradient. The codebook embeddings are updated using exponential moving averages (EMA) of their assigned encoder outputs.
We collect distillation rollouts from a multi-expert teacher by selecting the specialist expert associated with each motion category. For sequences without object observations, we pad the object-kinematic inputs with zeros.

\begin{figure}[!t]
\centering
\includegraphics[width=\columnwidth]{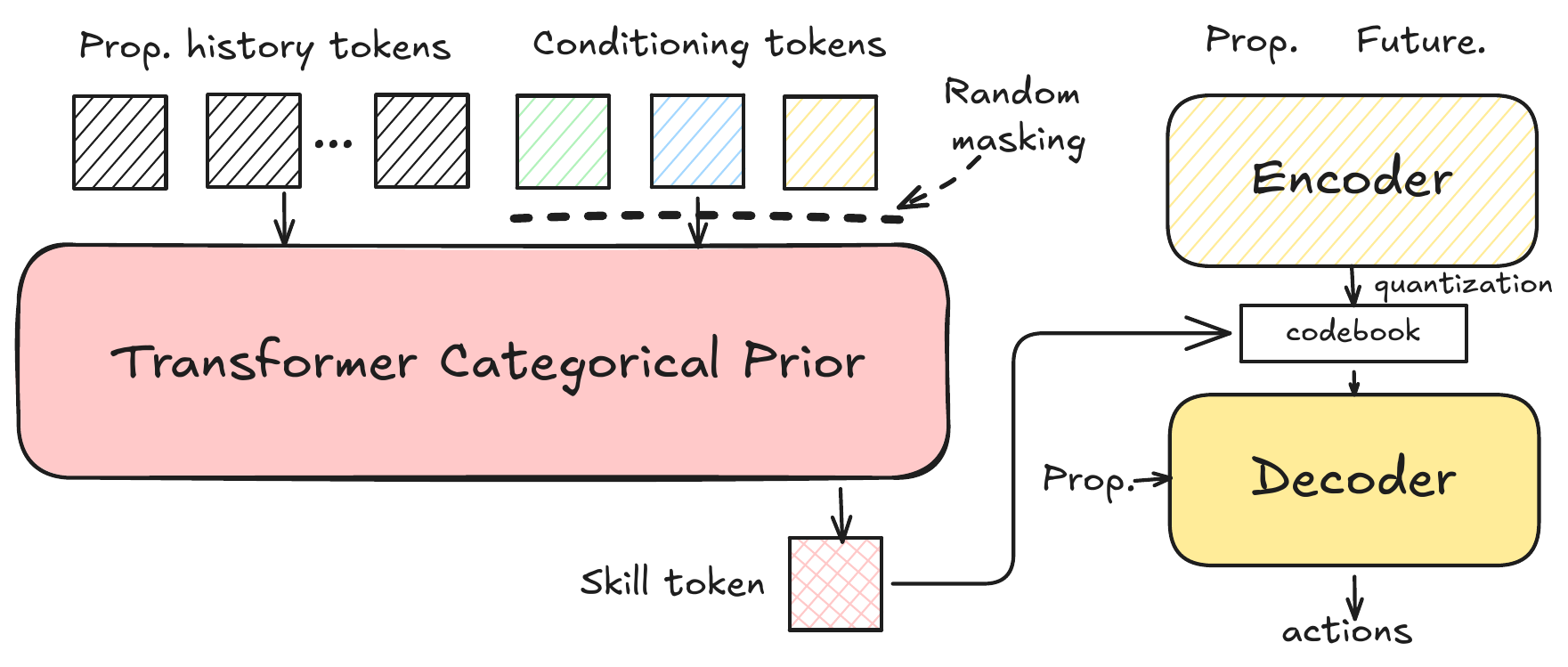}
\caption{\textbf{Architecture of the generative interaction prior.} Teacher behaviors are first encoded into discrete skill tokens, while multimodal interaction contexts and task goals are tokenized through a unified conditioning interface. A transformer-based prior predicts context-conditioned skill tokens, which are decoded into low-level humanoid actions.}
\Description{Architecture diagram in which proprioceptive-history tokens and randomly masked conditioning tokens enter a transformer categorical prior that predicts a discrete skill token. During training, an encoder quantizes teacher motion features through a codebook; the selected codebook embedding and current proprioception are passed to a decoder that outputs humanoid actions.}
\label{fig:generative_prior_arch}
\end{figure}

\paragraph{Unified Conditioning Interface.}
To model skill selection under heterogeneous interaction contexts and task goals, we introduce a unified conditioning interface. Scene context is represented by a local heightmap~\cite{peng2016terrain}, while object context is encoded by an object Basis Point Set (BPS)~\cite{prokudin2019efficient} and body-object proximity cues, i.e., per-joint distances to the object surface. We define the context and task-goal modalities as
\begin{equation}
\mathcal{M}_{C}=\{\text{height},\text{bps},\text{prox}\}, \qquad
\mathcal{M}_{G}=\{\text{human},\text{object}\}.
\end{equation}
When paired text descriptions are available, they can optionally be encoded as CLIP-based goal features and appended to the conditioning sequence using the same availability-mask mechanism.
For each modality $m\in\mathcal{M}_{C}\cup\mathcal{M}_{G}$, a modality-specific tokenizer maps the corresponding signal $x_{m,t}^j$ into a shared embedding space. We organize these embeddings into a unified conditioning sequence
\begin{equation}
\mathcal{U}_t^j =
\left[
\mathrm{Tok}_m(x_{m,t}^j)
\right]_{m\in\mathcal{M}_{C}\cup\mathcal{M}_{G}},
\end{equation}
and use a binary availability mask $\delta_m^j$ to exclude modalities absent from each data source. This shared interface allows \ourmethod to unify heterogeneous HSI and HOI conditioning within a single interaction prior.

\paragraph{Context-Conditioned Skill Prior.}
Building on the discrete skill representation, we train a transformer-based interaction prior $p_{\theta}$ that predicts a categorical distribution over skill tokens from the proprioceptive history and unified interaction context, $p_{\theta}(\cdot\mid s_{t-H:t}^{\mathrm{prop}},\mathcal{U}_t)$. With the teacher token $q_t^{E}=Q(E_{\eta}(o_t^E))$, the prior is trained using
\begin{equation}
\mathcal{L}_{\mathrm{prior}}
= -\sum_t \log p_{\theta}(q_t^{E} \mid s_{t-H:t}^{\mathrm{prop}}, \mathcal{U}_t).
\end{equation}

During training, we randomly mask available context and task-goal modalities with probability $p_{\mathrm{mask}}$, improving robustness to partial conditioning and enabling conditional guidance during downstream adaptation. The resulting prior models a context-dependent distribution over reusable skill tokens, grounding skill selection in scene, object, and task information.

\subsection{Stage III: Prior-Guided Online RL Post-training}

Although the distilled interaction prior captures diverse behaviors conditioned on environmental context and task semantics, downstream tasks may introduce objectives not fully specified by the pretraining data. We initialize a downstream task policy $\pi_{\phi}$ from the pretrained prior $p_{\theta}$ and adapt it with proximal policy optimization (PPO), while keeping the pretrained prior, VQ decoder $D_{\psi}$, and codebook embeddings $\mathcal{Z}$ frozen. The task policy is regularized toward a CFG-guided reference prior to preserve the learned interaction structure. We minimize the post-training loss $\mathcal{L}_{\mathrm{post}}=\mathcal{L}_{\mathrm{PPO}}+\lambda_{\mathrm{KL}}\mathcal{L}_{\mathrm{KL}}$, where
\begin{equation}
\mathcal{L}_{\mathrm{KL}}
=
\mathbb{E}_{t}\!\left[
D_{\mathrm{KL}}\!\left(
\pi_{\phi}(\cdot \mid s_{t-H:t}^{\mathrm{prop}}, \mathcal{U}_t)
\,\|\, 
p_{\theta}^{\mathrm{cfg}}(\cdot \mid s_{t-H:t}^{\mathrm{prop}}, \mathcal{U}_t)
\right)
\right].
\end{equation}

\paragraph{CFG-Guided Prior Regularization.}
Because the interaction prior must cover diverse behaviors across scenes, objects, and goals, its distribution can remain overly broad for a specific downstream objective. We therefore construct a more task-focused reference distribution using classifier-free guidance (CFG). During prior training, modality dropout enables the same prior to be queried unconditionally or with individual conditioning modalities. Let $f_{\theta}$ denote the pre-softmax logits of the frozen prior, and let $\mathcal{U}_{t,m}$ retain only modality $m$ while masking all others. During post-training, we compute
\begin{equation}
\ell_t^{\mathrm{uncond}}
=f_{\theta}(s_{t-H:t}^{\mathrm{prop}},\varnothing), \quad
\ell_t^m
=f_{\theta}(s_{t-H:t}^{\mathrm{prop}},\mathcal{U}_{t,m}),
\end{equation}
for each active modality $m$, and combine them as
\begin{equation}
\ell_t^{\mathrm{cfg}}
=
\ell_t^{\mathrm{uncond}}
+
\sum_{m \in \mathcal{A}(\mathcal{U}_t)}
w_m
\left(
\ell_t^m-\ell_t^{\mathrm{uncond}}
\right).
\end{equation}
The guided reference prior is obtained as $p_{\theta}^{\mathrm{cfg}}(\cdot\mid s_{t-H:t}^{\mathrm{prop}},\mathcal{U}_t)=\mathrm{Softmax}(\ell_t^{\mathrm{cfg}})$, where $\mathcal{A}(\mathcal{U}_t)$ denotes the active conditioning modalities and $w_m$ controls the guidance strength of modality $m$. This guidance provides a more task-focused reference while retaining the behavioral support of the pretrained interaction prior.
\begin{figure*}[!t]
\centering
\includegraphics[width=1.0\textwidth]{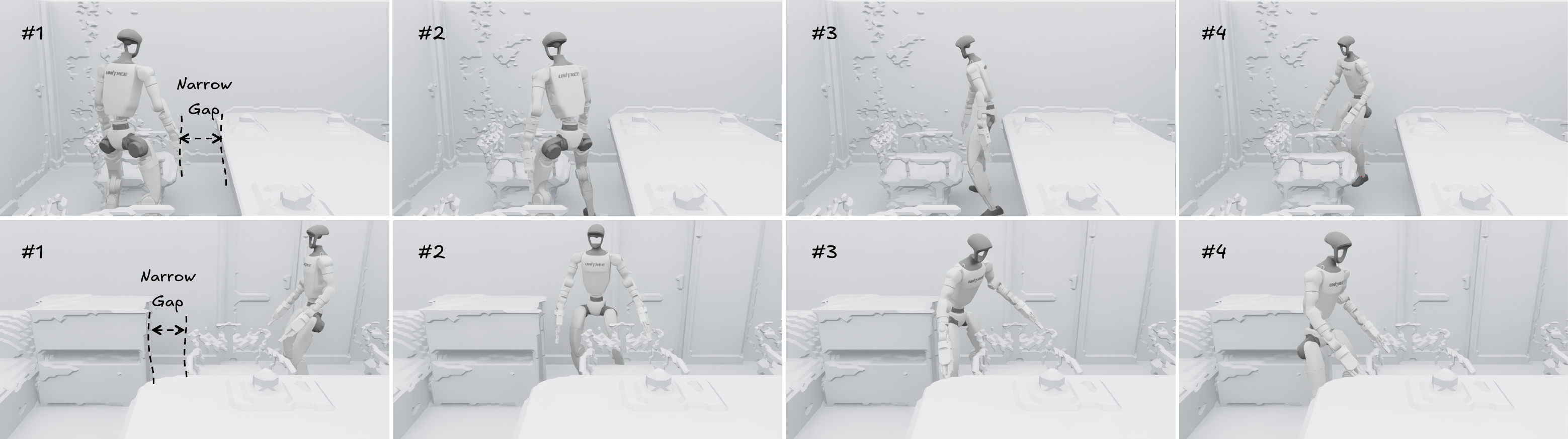}
\caption{\textbf{Qualitative results on scene navigation.} Representative snapshots show coordinated whole-body motion in cluttered scenes, including torso rotation and sideways stepping to pass through narrow gaps while maintaining stable, goal-directed locomotion.}
\Description{Two four-frame navigation sequences viewed from different angles. In each sequence, the humanoid rotates its torso and steps sideways to pass through a narrow gap between pieces of furniture without colliding with them.}
\label{fig:qualitative_collision_avoidance}
\end{figure*}

\begin{figure*}[!t]
\centering
\includegraphics[width=\textwidth]{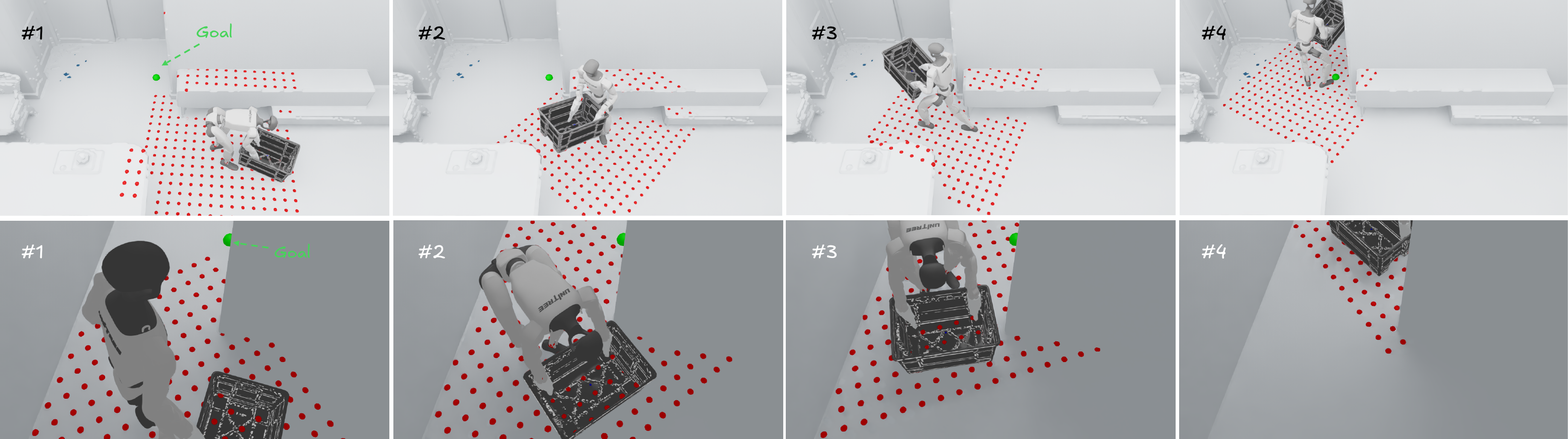}
\caption{\textbf{Qualitative results on compositional capabilities.} Representative snapshots show compositional behaviors acquired through RL post-training: the humanoid transports an object while remaining aware of the surrounding environment, bypassing obstacles by adapting its whole-body motion.}
\Description{Two four-frame sequences of a humanoid transporting a large crate through cluttered rooms toward a green goal marker. Red height-scan points indicate nearby scene geometry as the humanoid changes its body orientation and carrying posture to move around obstacles.}
\label{fig:composite_capabilities}
\end{figure*}

\section{Experiments}

\subsection{Experimental Setup}
\paragraph{Tasks.}
We instantiate three evaluation tasks: \textit{Scene Navigation}, \textit{Object Relocation}, and \textit{Compositional Task}. In \textit{Scene Navigation}, the humanoid navigates to a target location randomly sampled $1.5$--$3.0\,\mathrm{m}$ from its initial position while avoiding cluttered scene geometry; a rollout is successful if the humanoid base is within $0.15\,\mathrm{m}$ of the target in the $xy$ plane. In \textit{Object Relocation}, the humanoid moves an object to a target location sampled $1.0$--$2.0\,\mathrm{m}$ away through whole-body interaction; success is achieved when the object center is within $0.15\,\mathrm{m}$ of the target in the $xy$ plane. In \textit{Compositional Task}, the humanoid first navigates through the scene and then relocates the object within a single continuous episode, requiring a transition from HSI to HOI behaviors. A rollout is counted as successful only when both subgoals are completed.
The transition from navigation to object relocation is triggered by a scripted state machine once the navigation success condition is met.

\paragraph{Datasets.}
The training corpus combines public and self-collected motion data. We use TRUMANS~\cite{jiang2024scaling} for human-scene interaction (HSI) and OMOMO~\cite{li2023object} for human-object interaction (HOI), together with additional locomotion and box-transport motions. For OMOMO, we use motions that do not require dexterous hand grasping, matching our focus on whole-body humanoid control and object transport.

\paragraph{Evaluation Protocol.}
The main comparison in Table~\ref{tab:main_results} uses randomized in-distribution evaluation over all scenes and object instances used for downstream training, with initial states and goal commands independently resampled for each rollout. To evaluate scene generalization, we additionally test CHIP on ten scenes excluded from all training stages, while keeping the object-instance pool and task/goal distributions unchanged. For this held-out-scene evaluation, each scene--task pair is tested over 100 episodes. We do not evaluate generalization to held-out object instances.

\paragraph{Metrics.}
We report success rate (Succ.) as the primary metric for single-stage tasks and full-task success rate for the compositional task. We also report final distance-to-goal (Dist.), measured in the $xy$ plane between the character base and the navigation target for scene navigation, and between the object center and the object target for object relocation and compositional tasks. 
To quantify collision-avoidance capability, we report collision score (Coll.): a frame is counted as a collision frame if any non-foot body of the humanoid contacts the scene with a contact force greater than $1.0\,\mathrm{N}$, and the score is the fraction of collision frames over the rollout.
Finally, we report failure rate (Fail.), defined as the percentage of rollouts terminated by falling, where falling is detected when the humanoid base height drops below $0.2\,\mathrm{m}$.

\paragraph{Baselines.}
We compare against two representative categories of prior methods that test the key alternatives to our framework. 
First, we include TokenHSI~\cite{pan2025tokenhsi} as a task-oriented unified-policy baseline. It integrates heterogeneous task specifications within a shared transformer policy and acquires supported behaviors through task-oriented training.
This comparison contrasts policy-level unification with our approach of first distilling reusable interaction knowledge into a shared prior before downstream adaptation.
Second, we adapt MaskedMimic~\cite{tessler2024maskedmimic} as a motion-centric generalist-control baseline evaluated zero-shot with respect to downstream task optimization. We train it with environment-context observations as conditioning, but do not perform task-specific RL adaptation. This comparison tests whether context-conditioned motion pretraining alone transfers directly to our downstream tasks.

\paragraph{Implementation Details.}
All experiments are conducted in IsaacLab~\cite{mittal2025isaaclab} with a Unitree G1 humanoid model containing 29 actuated joints. The simulator runs at 200 Hz and applies control actions at 50 Hz. Unless otherwise specified, all stages use the same embodiment, simulator configuration, and control interface.

We preprocess heterogeneous motion data by retargeting SMPL-X-format human motions to the G1 embodiment. We use GMR~\cite{ze2025gmr} for contact-free locomotion and OmniRetargeting~\cite{yang2025omniretarget} for contact-rich interactions. For HSI motions, we additionally precompute egocentric height scans in the humanoid heading frame as scene context, while kinematic targets are extracted from the retargeted sequences.

Training follows the three-stage pipeline described in Sec.~\ref{sec:method}. Stage I trains PPO-based motion-imitation experts using physics-based tracking objectives. In Stage II, we distill these experts into a unified discrete skill representation and a context-conditioned categorical prior over skill tokens. The VQ-VAE uses MLPs for both the encoder and decoder, together with a 1,024-entry codebook of 64-dimensional embeddings updated by EMA~\cite{van2017neural}. The interaction prior is implemented as a transformer and optimized with cross-entropy over the discrete skill tokens. Skill tokens are predicted at a frequency of 50 Hz.

In Stage III, we optimize the task policy with PPO under prior regularization. At control time, the task policy predicts a categorical distribution over skill tokens at 50 Hz, from which a token is sampled and decoded into joint-space PD targets. We use a sampling temperature of $0.05$ during inference.

\subsection{Results on Unified Humanoid Interaction}
\begin{table*}[t]
\centering
\caption{\textbf{Quantitative results on heterogeneous and compositional humanoid interaction tasks.}
We report success and failure rates (\%), final distance-to-goal (m), and collision-frame fractions for Scene Navigation, Object Relocation, and the Compositional Task.}
\label{tab:main_results}
\resizebox{\textwidth}{!}{%
\begin{tabular}{l cccc ccc cccc} 
\toprule
& \multicolumn{4}{c}{Scene Nav.} & \multicolumn{3}{c}{Object Reloc.} & \multicolumn{4}{c}{Compositional} \\
\cmidrule(lr){2-5} \cmidrule(lr){6-8} \cmidrule(lr){9-12}
Method & Succ. $\uparrow$ & Dist. $\downarrow$ & Coll. $\downarrow$ & Fail. $\downarrow$ & Succ. $\uparrow$ & Dist. $\downarrow$ & Fail. $\downarrow$ & Succ. $\uparrow$ & Dist. $\downarrow$ & Coll. $\downarrow$ & Fail. $\downarrow$ \\
\midrule
TokenHSI~\cite{pan2025tokenhsi} & \textbf{93.3} & \textbf{0.05} & 0.03 & 1.6 & 84.8 & 0.14 & 1.2 & 66.8 & 0.19 & 0.04 & 1.8\\
MaskedMimic~\cite{tessler2024maskedmimic} & 33.8 & 0.36 & 0.13 & 5.1 & 44.1 & 0.36 & 8.4 & 0.0 & 0.52 & 0.15 & 14.1 \\
\midrule
\rowcolor[gray]{0.9}
\textbf{Ours} & 90.1 & \textbf{0.05} & \textbf{0.02} & \textbf{1.5} & \textbf{88.2} & \textbf{0.10} & \textbf{1.0} & \textbf{71.4} & \textbf{0.13} & \textbf{0.02} & \textbf{1.6} \\
\bottomrule
\end{tabular}%
}
\end{table*}

\paragraph{Scene-aware Navigation.}
We first evaluate whether explicit human-scene interaction (HSI) modeling enables whole-body control and navigation in cluttered environments. 
As shown in Table~\ref{tab:main_results}, \ourmethod achieves the lowest collision and failure rates on scene navigation while remaining competitive with the strongest baseline in success rate. 
This indicates that the controller does not simply optimize goal reaching, but preserves the whole-body coordination and scene awareness encoded in the data prior. TokenHSI attains the highest navigation success through task-oriented optimization, but exhibits a higher collision rate than \ourmethod.
MaskedMimic is trained with environment-context conditioning but evaluated without task-specific adaptation. Its lower success and more frequent scene contacts indicate that context-conditioned motion pretraining alone does not reliably transfer to goal-directed collision avoidance in this setting.
Figure~\ref{fig:qualitative_collision_avoidance} further shows coordinated collision-avoidance behaviors: the humanoid passes through narrow spaces, turns its body sideways, and adapts its posture to avoid obstacles while maintaining stable locomotion.

\begin{figure}[t]
\centering
\includegraphics[width=\columnwidth]{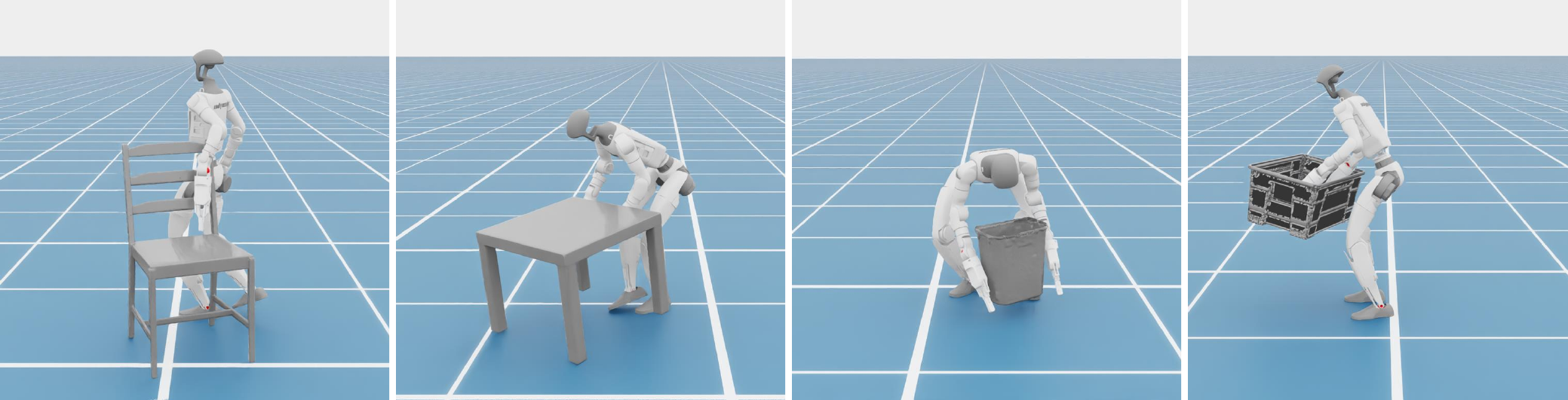}
\caption{\textbf{Diverse object interactions.} Qualitative results across the evaluated object instances and interaction types demonstrate coherent whole-body control.}
\Description{Four examples of whole-body object interaction on a blue grid floor. From left to right, the humanoid interacts with a chair, bends over a table, handles a large bin, and carries a rectangular crate.}
\label{fig:obj_manipulation}
\end{figure}

\paragraph{Object-aware Manipulation.}
We then evaluate whether the same framework can capture contact-rich interactions with objects. On object relocation, \ourmethod achieves the best success rate and lowest final distance in Table~\ref{tab:main_results}, demonstrating that the unified prior supports whole-body object transport in addition to scene-aware locomotion. In its zero-shot setting, MaskedMimic achieves substantially lower success and larger object-target errors, indicating limited transfer of motion-centric context-conditioned pretraining to the precise contact-rich objective without task-specific adaptation. TokenHSI can optimize task-specific behaviors, but it requires task-oriented training for each supported interaction setting, making extension to new object-centric tasks costly. Figure~\ref{fig:obj_manipulation} further illustrates coherent whole-body motion across the evaluated object instances and interaction types. These results suggest that the object-aware conditioning interface provides a reusable representation for HOI behaviors without requiring a separate downstream low-level controller for each object category.

\begin{figure}[t]
\centering
\includegraphics[width=\columnwidth]{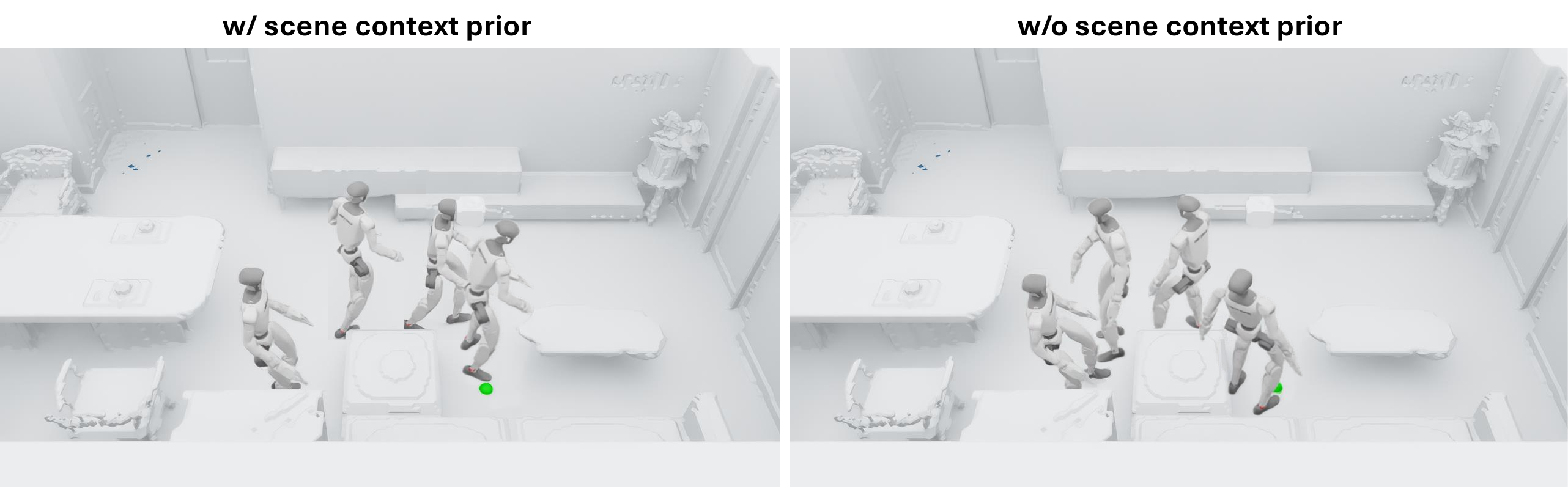}
\caption{\textbf{Ablation on scene-context conditioning.} Without scene-context conditioning, RL can exploit the task rewards and learn irregular behaviors, such as moving backward, instead of maintaining a natural walking gait in cluttered environments.}
\Description{Side-by-side multi-exposure trajectories in the same furnished room. With the scene-context prior, the humanoid walks naturally around furniture toward a green goal; without it, the humanoid reaches the goal using an irregular backward-facing trajectory.}
\label{fig:ablation_scenecontext}
\end{figure}

\paragraph{Compositional Capabilities.}
Finally, we evaluate whether \ourmethod can execute multiple interaction capabilities sequentially within a continuous episode. According to Table~\ref{tab:main_results}, our approach outperforms the baselines in full-task success, final distance, collision rate, and failure rate. Qualitatively, Figure~\ref{fig:teaser} shows that the policy maintains continuous control with smooth transitions from locomotion to manipulation, suggesting that the shared prior provides a common interaction space for connecting heterogeneous behaviors.

Beyond sequential chaining, \ourmethod also supports concurrent capability composition, where the humanoid must maintain environmental awareness while simultaneously manipulating an object. As shown in Figure~\ref{fig:composite_capabilities}, the humanoid can manipulate an object while avoiding obstacles and continuously adjusting its whole-body posture throughout the rollout.
Through unified modeling of interaction contexts, RL post-training can acquire such compositional capabilities even when they are not directly present as complete trajectories in the data.

\paragraph{Held-out Scene Generalization.}
We further evaluate CHIP on ten scenes excluded from all training stages. As shown in Table~\ref{tab:heldout_scene_results}, CHIP maintains strong performance on unseen scene geometry. Object instances and task and goal distributions remain unchanged.
\begin{table}[t]
  \centering
  \caption{\textbf{Held-out scene generalization of CHIP.}
  We compare randomized in-distribution evaluation with ten scenes excluded from all training stages, while keeping the object-instance pool and task and goal distributions unchanged.}
  \label{tab:heldout_scene_results}
  \resizebox{\columnwidth}{!}{\begin{tabular}{l cc cc}
      \toprule
      Setting & Nav. Succ. $\uparrow$ & Nav. Fail. $\downarrow$ &
      Comp. Succ. $\uparrow$ & Comp. Fail. $\downarrow$ \\
      \midrule
      In-distribution  & 91.3 & 1.8 & 69.8 & 4.8 \\
      Held-out scenes & 86.1 & 5.5 & 60.2 & 10.5 \\
      \bottomrule
    \end{tabular}}
\end{table}

\subsection{Ablation Study}
\newcommand{\cmark}{\ding{51}}
\newcommand{\xmark}{\ding{55}}

\begin{table}[t]
\centering
\caption{\textbf{Component-wise ablation of~\ourmethod.} We report success rates on downstream tasks to evaluate the contributions of RL post-training, prior regularization, and context-conditioned interaction prior modeling.}
\label{tab:matrix_ablation}
\resizebox{\columnwidth}{!}{%
\begin{tabular}{l ccc | ccc}
\toprule
\makecell[l]{Variant}
& \makecell{RL\\Post-train.}
& \makecell{Prior Reg.\\Loss}
& \makecell{Inter.\\Context}
& \makecell{Scene\\Nav.}
& \makecell{Object\\Reloc.}
& \makecell{Comp.\\Task} \\
\midrule
Pretrained prior only & \xmark & \xmark & \cmark & 48.9 & 55.2 & 4.8 \\
\midrule
w/o prior reg. & \cmark & \xmark & \xmark & \textbf{94.2} & 0.0 & 0.0 \\
w/o context prior & \cmark & \cmark & \xmark & 92.1 & 0.0 & 0.0 \\
\rowcolor[gray]{0.9} \textbf{Full} & \cmark & \cmark & \cmark & 90.1 & \textbf{88.2} & \textbf{71.4} \\
\bottomrule
\end{tabular}%
}
\end{table}

\paragraph{RL Post-training.}
We first evaluate the contribution of Stage III by comparing the pretrained prior before and after RL post-training. The pretrained prior already provides broad behavior coverage from heterogeneous interaction data, but its performance remains limited on downstream objectives that are not fully specified during pretraining. RL post-training substantially improves task success, especially for compositional settings, indicating that this stage specializes the shared prior to new objectives rather than relearning interaction behaviors from scratch.

\paragraph{Prior Regularization.}
We further ablate prior regularization during post-training to study whether preserving the pretrained interaction prior benefits downstream adaptation. As shown in Table~\ref{tab:matrix_ablation}, the skill prior substantially facilitates learning contact-rich object picking and manipulation. These interactions require precise whole-body coordination and are difficult to acquire from sparse or weak task rewards alone.

\paragraph{Context-Conditioned Prior.}
Finally, we investigate the role of the context-conditioned interaction prior by replacing it with an unconditional prior for regularization. 
This ablation tests whether explicitly modeling interaction context is more effective than using a context-agnostic prior. Without scene and object conditioning, the regularization prior does not encode how motions should adapt to surrounding geometry, object state, or task goals. As shown in Table~\ref{tab:matrix_ablation}, the controller struggles to acquire object-manipulation skills without prior knowledge of object interaction. 
It also fails to capture the spatial relationship between motion and the surrounding environment, leading to unnatural behaviors in cluttered scenes, as illustrated in Figure~\ref{fig:ablation_scenecontext}.

\section{Conclusion}
We presented \ourmethod, a unified framework for physics-based humanoid control across diverse interaction tasks in complex 3D environments. By learning a multimodal context-conditioned interaction prior from heterogeneous data, \ourmethod captures reusable scene-aware and object-aware behaviors within a shared skill space. Policies adapted from this prior support whole-body collision avoidance, contact-rich object manipulation, and compositional behaviors such as environment-aware object transport and multi-stage execution of composed interaction skills. Our three-stage pipeline consolidates specialist behaviors into a reusable interaction prior while supporting smooth transitions, generalization to unseen scenes, and downstream adaptation.

Several limitations remain. The breadth of interaction grounding in \ourmethod depends on the quality and coverage of motion data paired with scene or object context, which remains limited. In addition, the current policy uses relatively coarse geometric observations and does not yet fully exploit semantic scene or object understanding. Future work could replace the current scripted high-level task sequencing with learned planning, while incorporating richer perceptual inputs such as RGB or depth observations to support more autonomous and semantically grounded interactions in diverse environments.

\begin{acks}
  This research was supported in part by the Australian Government through the Australian Research Council (ARC) Discovery Project DP260100218.
\end{acks}

\bibliographystyle{ACM-Reference-Format}
\bibliography{reference}


\clearpage
\appendix

This supplementary material provides additional details about the datasets, downstream evaluation tasks, and implementation settings used in our experiments.

\section{Data Statistics}
\label{sec:supp_data_statistics}

\subsection{Dataset Composition}
We train \ourmethod on a heterogeneous motion corpus consisting of standalone locomotion, box-transport, human-scene interaction (HSI), and human-object interaction (HOI) data. After applying the training-time motion filters, the final training set contains 2,826 motion sequences with a total duration of 12.17 hours.
The corpus comprises four components: standalone locomotion sequences, box-transport sequences, a subset of \texttt{TRUMANS}, and a subset of \texttt{OMOMO}. Sequence duration is computed from the original frame count and source FPS, and aggregate durations are computed before rounding the per-source statistics.

\begin{table}[h]
\centering
\caption{\textbf{Motion data statistics.} We report the number of files, training sampling weight, sequence duration, and scene/object availability for each data source after applying the training-time motion filters.}
\label{tab:supp_data_statistics}
\resizebox{\columnwidth}{!}{%
\begin{tabular}{lrrrccr}
\toprule
Data source & Files & Weight & Avg. Len. (s) & \makecell{Has\\Object} & \makecell{Has\\Scene} & Total Len. (h) \\
\midrule
\texttt{locomotion} & 6 & 0.05 & 389.06 & No & No & 0.65 \\
\texttt{box transport} & 120 & 0.05 & 22.00 & Yes & No & 0.73 \\
\texttt{TRUMANS (idle)} & 1486 & 0.45 & 21.11 & No & Yes & 8.71 \\
\texttt{OMOMO (subset)} & 1214 & 0.45 & 6.14 & Yes & No & 2.07 \\
\midrule
Total & 2826 & 1.00 & 15.50 & -- & -- & 12.17 \\
\bottomrule
\end{tabular}%
}
\end{table}

Table~\ref{tab:supp_data_statistics} summarizes the composition of the pretraining data. The standalone locomotion and box-transport components contribute 126 sequences and 1.38 hours in total. TRUMANS contributes 1,486 HSI sequences and 8.71 hours, while OMOMO contributes 1,214 HOI sequences and 2.07 hours. OMOMO also provides text annotations and CLIP text embeddings, which are included as an optional goal-conditioning modality during prior training. The downstream experiments do not report a separate evaluation of text conditioning.

\subsection{Expert Assignment and Conditioning Modalities}
We further group the pretraining motions by expert assignment. The object-centric experts correspond to \texttt{trashcan}, \texttt{plasticbox}, \texttt{largetable}, and \texttt{woodchair}, containing 317, 300, 237, and 360 sequences, respectively. The remaining experts cover locomotion and box transport (126 sequences) and HSI motions from \texttt{TRUMANS (idle)} (1,486 sequences). In total, 1,214 sequences provide optional text annotations and CLIP text embeddings.

\section{Implementation Details}

\subsection{Training Configurations}
Tables~\ref{tab:supp_stage1_hyperparameters}--\ref{tab:supp_stage3_hyperparameters} summarize the architectures and optimization settings used in the three training stages.

\begin{table}[H]
\centering
\small
\caption{\textbf{Key hyperparameters for Stage I motion imitation.}}
\label{tab:supp_stage1_hyperparameters}
\begin{tabular}{p{0.48\columnwidth} p{0.42\columnwidth}}
\toprule
Hyperparameter & Value \\
\midrule
RL algorithm & PPO \\
Optimizer & AdamW \\
Tracking actor & MLP $[512,256,128]$, ELU, tanh output \\
Tracking critic & MLP $[512,256,128]$, ELU, linear value output \\
Learning rate & $1\times10^{-3}$ \\
Learning-rate schedule & Adaptive \\
Weight decay & $1\times10^{-6}$ \\
Discount factor $\gamma$ & $0.99$ \\
GAE coefficient $\lambda$ & $0.95$ \\
PPO clipping coefficient $\epsilon$ & $0.2$ \\
Actor loss coefficient & $1.0$ \\
Critic loss coefficient & $1.0$ \\
Entropy coefficient & $0.005$ \\
PPO rollout horizon & $24$ \\
Minibatch size & $24{,}576$ \\
Mini-epochs & $5$ \\
Gradient clipping & $1.0$ \\
Reference-state initialization & Enabled \\
Kinematic early termination & Enabled \\
Future-reference input & One frame \\
Simulation / control frequency & $200/50\,\mathrm{Hz}$ \\
\bottomrule
\end{tabular}
\end{table}

\begin{table}[H]
\centering
\small
\caption{\textbf{Stage II skill-distillation and prior-learning hyperparameters.}}
\label{tab:supp_stage2_hyperparameters}
\begin{tabular}{@{}p{0.39\columnwidth} p{0.56\columnwidth}@{}}
\toprule
Hyperparameter & Value \\
\midrule
VQ-VAE encoder & MLP $[1024,1024,768,512]$; ReLU \\
VQ-VAE decoder & MLP $[1024,1024,768,512]$; ReLU \\
Codebook & $K=1024$; 64-D embeddings; one quantizer \\
Commitment weight $\beta_{\mathrm{com}}$ & $0.25$ \\
EMA decay & $0.99$ \\
Prior backbone & 4-layer Transformer; model dimension $512$; 8 attention heads \\
Prior FFN & Dimension $1536$; GELU; dropout $0$ \\
Prior-history input & 5 proprioceptive frames \\
Modality-masking probability $p_{\mathrm{mask}}$ & $0.3$ \\
Optimizer & AdamW \\
Learning rate & $2\times10^{-4}$ \\
Learning-rate schedule & Constant \\
Teacher data-collection horizon & $32$ \\
Minibatch size & $8{,}192$ \\
Mini-epochs & $3$ \\
Teacher data-collection environments & $4{,}096$ \\
Gradient clipping & $1.0$ \\
Skill-token frequency & $50\,\mathrm{Hz}$ \\
\bottomrule
\end{tabular}
\end{table}

\begin{table}[H]
\centering
\small
\caption{\textbf{Stage III RL post-training hyperparameters.}}
\label{tab:supp_stage3_hyperparameters}
\begin{tabular}{@{}p{0.39\columnwidth} p{0.56\columnwidth}@{}}
\toprule
Hyperparameter & Value \\
\midrule
RL algorithm & PPO \\
Actor backbone & 4-layer Transformer; model dimension $512$; 8 attention heads \\
Actor FFN / output & FFN dimension $1536$; GELU; dropout $0$; 1024-way categorical skill distribution \\
Actor / prior history input & 5 proprioceptive frames \\
Policy initialization & Pretrained interaction prior \\
Trainable modules & Task policy \\
Frozen modules & Interaction prior, VQ decoder, and codebook embeddings \\
Prior backbone & 4-layer Transformer; model dimension $512$; 8 attention heads; FFN dimension $1536$ \\
Codebook & $K=1024$; 64-D embeddings \\
Motion decoder & MLP $[1024,1024,768,512]$; ReLU \\
Critic inputs & Privileged state, target location, and local heightmap \\
Critic backbone & 2-layer Transformer; model dimension $512$; 4 attention heads; FFN dimension $1024$ \\
Critic head & MLP $[512,256,1]$ \\
Optimizer & AdamW \\
Learning rate & $5\times10^{-5}$ \\
Weight decay & $1\times10^{-6}$ \\
Learning-rate schedule & Constant \\
Discount factor $\gamma$ & $0.99$ \\
GAE coefficient $\lambda$ & $0.95$ \\
PPO clipping coefficient $\epsilon$ & $0.2$ \\
Rollout horizon & $32$ \\
Minibatch size & $8{,}192$ \\
Mini-epochs & $2$ \\
Number of environments & $4{,}096$ \\
Gradient clipping & $1.0$ \\
KL coefficient $\lambda_{\mathrm{KL}}$ & $0.01$ \\
CFG weight $w_m$ & $1.5$ for all active modalities \\
Inference sampling temperature & $0.05$ \\
Control / token frequency & $50/50\,\mathrm{Hz}$ \\
\bottomrule
\end{tabular}
\end{table}

\subsection{Stage I Motion-Imitation Objective}
Let $a$ denote the pelvis anchor, $\mathcal{B}$ the 14 equally weighted tracking bodies, and $\mathcal{O}$ the objects. Reference quantities are marked by $(\cdot)^\star$. All tracking terms use
\begin{equation}
\begin{aligned}
K_{\sigma}(e)&=\exp\!\left[-(e/\sigma)^2\right],\\
d_R(q_1,q_2)&=2\arccos\!\left(\operatorname{clip}(|q_1^\top q_2|,0,1)\right).
\end{aligned}
\end{equation}
The per-step reward is
\begin{equation}
\begin{aligned}
r_t^E=\Delta t\big(&0.5r_a^p+0.5r_a^R+r_B^p+r_B^R+r_B^v+r_B^\omega\\
&+0.5r_O^p+0.5r_O^R-0.1P_{\Delta u}-10P_{\mathrm{lim}}-0.1P_{\mathrm{contact}}\big),
\end{aligned}
\end{equation}
with control step $\Delta t=\texttt{decimation}\times\texttt{sim.dt}=4\times(1/200)=0.02\,\mathrm{s}$.

\begin{equation}
\begin{aligned}
e_a^p&=\left\|p_{a,t}^\star-p_{a,t}\right\|_2,
&r_a^p&=K_{\sigma_a^p}(e_a^p),\\
e_a^R&=d_R(q_{a,t}^\star,q_{a,t}),
&r_a^R&=K_{\sigma_a^R}(e_a^R).
\end{aligned}
\end{equation}

For body-pose tracking, the reference is aligned to the current pelvis in planar position and yaw while retaining the reference pelvis height:
\begin{equation}
q_{\Delta,t}=\operatorname{Yaw}\!\left(q_{a,t}\otimes(q_{a,t}^\star)^{-1}\right),
\qquad
\bar p_{a,t}=\begin{bmatrix}p_{a,t}^{x}&p_{a,t}^{y}&p_{a,t}^{\star,z}\end{bmatrix}^{\!\top},
\end{equation}
\begin{equation}
\widetilde p_{b,t}^\star
=\bar p_{a,t}+R(q_{\Delta,t})(p_{b,t}^\star-p_{a,t}^\star),
\qquad
\widetilde q_{b,t}^\star=q_{\Delta,t}\otimes q_{b,t}^\star.
\end{equation}
\begin{equation}
\begin{aligned}
e_b^p&=\|\widetilde p_{b,t}^\star-p_{b,t}\|_2,
&e_b^R&=d_R(\widetilde q_{b,t}^\star,q_{b,t}),\\
e_b^v&=\left\|v_{b,t}^\star-v_{b,t}\right\|_2,
&e_b^\omega&=\left\|\omega_{b,t}^\star-\omega_{b,t}\right\|_2.
\end{aligned}
\end{equation}
\begin{equation}
r_B^x=\frac{1}{14}\sum_{b\in\mathcal{B}}K_{\sigma_B^x}(e_b^x),
\qquad x\in\{p,R,v,\omega\}.
\end{equation}

Let $m_{o,t}\in\{0,1\}$ indicate whether object $o$ is active in the current reference frame. We track its position relative to the anchor and its global orientation using
\begin{equation}
e_o^p=\left\|(p_{o,t}^\star-p_{a,t}^\star)-(p_{o,t}-p_{a,t})\right\|_2,
\qquad
e_o^R=d_R(q_{o,t}^\star,q_{o,t}).
\end{equation}
\begin{equation}
r_O^x=
\frac{\sum_{o\in\mathcal{O}}m_{o,t}K_{\sigma_O^x}(e_o^x)}
{\sum_{o\in\mathcal{O}}m_{o,t}+10^{-6}},
\qquad x\in\{p,R\},
\end{equation}
which is zero when no object is active.

\begin{equation}
\begin{aligned}
P_{\Delta u}&=\|u_t-u_{t-1}\|_2^2,\\
P_{\mathrm{lim}}&=\sum_j\min\!\left(
[q_j^{\mathrm{low}}-q_{j,t}]_+ +[q_{j,t}-q_j^{\mathrm{high}}]_+,1\right),\\
P_{\mathrm{contact}}&=\sum_{\ell\in\mathcal{C}_{\mathrm{undesired}}}
\mathbb{I}\!\left[\max_{\tau\in\mathcal{H}}\|F_{\ell,\tau}\|_2>1\,\mathrm{N}\right].
\end{aligned}
\end{equation}
Here $[x]_+=\max(x,0)$. Wrist and ankle contacts are permitted; all other body links belong to $\mathcal{C}_{\mathrm{undesired}}$.

\begin{table}[t]
\centering
\caption{\textbf{Stage I reward weights and fixed RBF bandwidths.}}
\label{tab:supp_tracking_rewards}
\resizebox{\columnwidth}{!}{%
\begin{tabular}{lcc}
\toprule
Reward component & Weight & $\sigma$ \\
\midrule
Anchor position & $0.5$ & $0.30\,\mathrm{m}$ \\
Anchor orientation & $0.5$ & $0.40\,\mathrm{rad}$ \\
Relative body position & $1.0$ & $0.30\,\mathrm{m}$ \\
Relative body orientation & $1.0$ & $0.40\,\mathrm{rad}$ \\
Global body linear velocity & $1.0$ & $1.00\,\mathrm{m/s}$ \\
Global body angular velocity & $1.0$ & $3.14\,\mathrm{rad/s}$ \\
Object relative position & $0.5$ & $0.30\,\mathrm{m}$ \\
Object global orientation & $0.5$ & $0.40\,\mathrm{rad}$ \\
Action rate & $-0.1$ & -- \\
Joint limit & $-10.0$ & -- \\
Undesired contacts & $-0.1$ & -- \\
\bottomrule
\end{tabular}%
}
\end{table}

We use fixed bandwidths in all reported experiments; adaptive bandwidth updates are disabled.

\subsection{Stage II Skill-Distillation and Prior-Learning Objectives}
For Stage II optimization, we use the straight-through quantized representation
\begin{equation}
\tilde z_t=z_t+\operatorname{sg}\!\left[z^{(q_t)}-z_t\right],
\end{equation}
which selects $z^{(q_t)}$ in the forward pass while passing decoder gradients directly to the encoder. The behavior-cloning term is
\begin{equation}
\mathcal{L}_{\mathrm{bc}}=-\sum_t\log D_{\psi}(a_t^E\mid s_t^{\mathrm{prop}},\tilde z_t).
\end{equation}
The complete skill-distillation objective is
\begin{equation}
\mathcal{L}_{\mathrm{VQ}}
=\mathcal{L}_{\mathrm{bc}}
+\beta_{\mathrm{com}}\sum_t
\left\|z_t-\operatorname{sg}\!\left[z^{(q_t)}\right]\right\|_2^2,
\end{equation}
with $\beta_{\mathrm{com}}=0.25$. The codebook is updated by exponential moving averages with decay $0.99$.

For prior learning, the proprioceptive history contains five frames. The teacher token is $q_t^E=Q(E_{\eta}(o_t^E))$, and the categorical interaction prior is trained with
\begin{equation}
\mathcal{L}_{\mathrm{prior}}
=-\sum_t\log p_{\theta}
\!\left(q_t^E\mid s_{t-H:t}^{\mathrm{prop}},\mathcal{U}_t\right).
\end{equation}
During training, we independently mask each available context or task-goal modality. The masking probability is $p_{\mathrm{mask}}=0.3$.

\subsection{Baseline Implementations}
TokenHSI~\cite{pan2025tokenhsi} serves as a task-oriented unified-policy baseline. We train its shared transformer policy using the supported downstream task specifications and task-oriented objectives.

We adapt MaskedMimic~\cite{tessler2024maskedmimic} as a motion-centric generalist-control baseline. Interaction-context and goal-conditioning signals are provided during training, but no task-specific RL adaptation is performed. It is therefore evaluated zero-shot with respect to downstream task optimization, testing whether context-conditioned motion pretraining alone transfers directly to our tasks.

Both baselines use the same training motion corpus, goal-conditioning signals, and interaction-context observations as \ourmethod. They also use the same humanoid embodiment, simulation and control frequencies, randomized evaluation protocol, and metric definitions.

\section{Downstream Tasks}
\label{sec:supp_downstream_tasks}

We provide additional details for the downstream tasks used during RL post-training. Each task combines sparse success or failure signals with task-specific shaping rewards and a shared set of control regularization terms. The shared penalties include action-rate smoothness, joint-limit violation, undesired contacts, and energy consumption. Table~\ref{tab:supp_task_obs_rewards} provides an overview of the task-specific observations and reward weights; the following subsections define their construction and usage in detail.

\begin{table*}[!b]
\centering
\caption{\textbf{Observation components and reward weights for downstream tasks.} We list task-specific observations and the reward weights used during RL post-training. Common penalties are shared across tasks.}
\label{tab:supp_task_obs_rewards}
\resizebox{\textwidth}{!}{%
\begin{tabular}{p{0.16\textwidth} p{0.38\textwidth} p{0.38\textwidth}}
\toprule
Task & Observation components & Reward weights \\
\midrule
Scene Navigation &
Local $16\times16$ heightmap covering $1.6\,\mathrm{m}\times1.6\,\mathrm{m}$; planar target displacement in the humanoid heading frame, stored in a shared padded 10-D target tensor. &
$w_{\mathrm{pos}}=4.0$, $w_{\mathrm{improve}}=2.0$, $w_{\mathrm{collision}}=-10.0$, $w_{\mathrm{vel}}=-10.0$, $w_{\mathrm{success}}=50.0$, $w_{\mathrm{failure}}=-50.0$, $w_{\mathrm{still}}=2.0$; common penalties: $w_{\mathrm{action\_rate}}=-0.01$, $w_{\mathrm{energy}}=-0.0001$, $w_{\mathrm{joint\_limit}}=-0.1$, $w_{\mathrm{contact}}=-0.1$. \\
\midrule
Object Relocation &
Local $16\times16$ heightmap; heading-frame object goal; 96-D body--object interaction geometry; 512-D object BPS; optional normalized 512-D CLIP text feature. &
$w_{\mathrm{target}}=4.0$, $w_{\mathrm{improve}}=2.0$, $w_{\mathrm{success}}=20.0$, $w_{\mathrm{failure}}=-10.0$; common penalties: $w_{\mathrm{action\_rate}}=-0.01$, $w_{\mathrm{energy}}=-0.0001$, $w_{\mathrm{joint\_limit}}=-0.1$, $w_{\mathrm{contact}}=-0.1$. \\
\bottomrule
\end{tabular}%
}
\end{table*}

\subsection{Scene Navigation}
\paragraph{Observations.}
The task-level navigation observation consists of a local $16\times16$ heightmap and the planar displacement to the target. The heightmap records scene height relative to the humanoid over a $1.6\,\mathrm{m}\times1.6\,\mathrm{m}$ region aligned with the humanoid heading. The target displacement is expressed in the same heading frame and stored in a shared padded 10-D target tensor, with the remaining eight dimensions set to zero.

\paragraph{Rewards.}
Let $d_t^{\mathrm{nav}}$ denote the planar distance between the humanoid base and the navigation target at step $t$. The main goal-reaching term is
\begin{equation}
    r_{\mathrm{pos}}^{\mathrm{nav}} = \exp\left(-\frac{\max(d_t^{\mathrm{nav}} - d_{\mathrm{th}}^{\mathrm{nav}}, 0)}{0.5}\right),
\end{equation}
where $d_{\mathrm{th}}^{\mathrm{nav}}$ is the target-reaching threshold. We also use a distance-improvement reward to encourage progress toward the target:
\begin{equation}
    r_{\mathrm{improve}}^{\mathrm{nav}} = \mathrm{clip}(d_{t-1}^{\mathrm{nav}} - d_t^{\mathrm{nav}}, -1, 1),
\end{equation}
which is set to zero once the humanoid enters the target region. Safety-related terms penalize scene collisions and excessive root velocity. For this training penalty, a collision is detected from either geometric intersection with scene obstacles or physical contact forces on non-allowed body parts. Near the target, we additionally use a stillness reward that encourages the humanoid to slow down and maintain a stable standing posture:
\begin{equation}
    r_{\mathrm{still}}^{\mathrm{nav}} = \mathbb{I}[d_t^{\mathrm{nav}} < d_{\mathrm{th}}^{\mathrm{nav}}]
    \exp\left(-\frac{\|\mathbf{v}_{xy}\|}{\sigma_v}\right)
    \exp\left(-\frac{\|\mathbf{q} - \mathbf{q}_{\mathrm{default}}\|^2}{\sigma_q}\right).
\end{equation}

\subsection{Object Relocation}
\paragraph{Observations.}
The object-relocation policy receives the same local $16\times16$ heightmap used for scene navigation, together with object-centric geometry and goal features. To encode body--object proximity, we sample 256 points on the object surface and find the nearest surface point to each of the 32 robot bodies. Let $\mathbf{d}_{b,t}^{H}$ denote the resulting body-to-object vector expressed in the humanoid heading frame. The interaction feature is
\begin{equation}
\mathbf{I}_{b,t}^{\mathrm{obj}}
=
\frac{\mathbf{d}_{b,t}^{H}}
{\|\mathbf{d}_{b,t}^{H}\|_2+\epsilon}
\exp\!\left(-5\|\mathbf{d}_{b,t}^{H}\|_2\right),
\qquad
\mathbf{I}_t^{\mathrm{obj}}
=\operatorname{vec}\!\left(\{\mathbf{I}_{b,t}^{\mathrm{obj}}\}_{b=1}^{32}\right)
\in\mathbb{R}^{96}.
\end{equation}
This representation jointly captures the direction and distance from each body to the object, with distant objects producing features close to zero.

We additionally represent object geometry using a Basis Point Set (BPS). We transform 512 fixed unit-ball basis points $\boldsymbol{\beta}_m$ by the humanoid root pose and record their nearest distances to the object surface:
\begin{equation}
B_{m,t}^{\mathrm{obj}}
=\min\!\left(
\min_k\left\|
\mathbf{p}_t^r+R(q_t^r)\boldsymbol{\beta}_m-\mathbf{s}_{k,t}^{o}
\right\|_2,
10
\right),
\qquad
\mathbf{B}_t^{\mathrm{obj}}\in\mathbb{R}^{512}.
\end{equation}

The object goal is represented by its 3-D position relative to the humanoid in the heading frame,
\begin{equation}
\mathbf{g}_t^{\mathrm{goal}}
=R_z(\psi_t)^\top(\mathbf{p}_t^g-\mathbf{p}_t^r)
\in\mathbb{R}^{3}.
\end{equation}
It occupies the first three entries of the shared padded 10-D target tensor and is not distance-clipped. Thus, the observation specifies the goal relative to the humanoid rather than the object-to-goal displacement.

When text conditioning is enabled, we use an $L_2$-normalized OpenCLIP ViT-B/32 feature for prompt $\tau_t$,
\begin{equation}
\mathbf{c}_t
=\frac{f_{\mathrm{CLIP}}(\tau_t)}{\|f_{\mathrm{CLIP}}(\tau_t)\|_2}
\in\mathbb{R}^{512}.
\end{equation}
One of four action prompts describing pushing, lifting and moving, pulling, or kicking is sampled uniformly for each environment.

\paragraph{Rewards.}
Let $d_t^{\mathrm{obj}}=\|\mathbf{p}_{\mathrm{obj},t}-\mathbf{p}_{\mathrm{target}}\|$ denote the 3-D object-target distance. The primary shaping term is
\begin{equation}
    r_{\mathrm{target}}^{\mathrm{obj}} = \exp\left(-\frac{d_t^{\mathrm{obj}}}{0.5}\right).
\end{equation}
We also use a distance-improvement reward,
\begin{equation}
    r_{\mathrm{improve}}^{\mathrm{obj}} = \mathrm{clip}(d_{t-1}^{\mathrm{obj}} - d_t^{\mathrm{obj}}, -1, 1).
\end{equation}
During RL post-training, success termination is triggered when the 3-D object-target distance remains below $0.15\,\mathrm{m}$ for $2.5\,\mathrm{s}$. This training-time termination criterion is distinct from the $xy$-plane distance metric reported in the main paper. Failure is triggered when the humanoid base height falls below the termination threshold.

\subsection{Compositional Task}
\paragraph{State-machine transition.}
The compositional task uses the scripted state machine described in the main paper. During the navigation phase, the policy observes the local heightmap and the current object position as its planar target. The relative object displacement in the humanoid heading frame is clipped to a maximum norm of $4\,\mathrm{m}$:
\begin{equation}
\begin{aligned}
\mathbf{d}_{t,xy}^{\mathrm{obj}}
&=\left[R_z(\psi_t)^\top(\mathbf{p}_t^o-\mathbf{p}_t^r)\right]_{xy},
\\
\overline{\mathbf{d}}_{t,xy}^{\mathrm{obj}}
&=\mathbf{d}_{t,xy}^{\mathrm{obj}}
\min\!\left(1,\frac{4}{\|\mathbf{d}_{t,xy}^{\mathrm{obj}}\|_2+\epsilon}\right).
\end{aligned}
\end{equation}
The clipped 2-D displacement occupies the first two entries of the padded 10-D target tensor. Interaction geometry, BPS, and text features are masked during this phase. Once navigation succeeds, the target tensor switches to the 3-D object goal, and the interaction geometry, BPS, and CLIP features described above are activated. The corresponding observations and rewards are therefore replaced by those of object relocation, and the rollout succeeds only after both stages are completed in sequence. Only the stage transition is scripted; control in both stages is produced by the learned task policy.

As in the main experiments, each raw reward term is multiplied by its corresponding weight and the simulation step duration during optimization.

\end{document}